\documentclass[letterpaper, 10 pt, conference]{inc/ieeeconf}
\IEEEoverridecommandlockouts                             
\usepackage{subcaption}
\usepackage{xcolor}
\usepackage{comment}
\usepackage{multirow}
\usepackage{colortbl}
\usepackage{array}
\usepackage{booktabs}
\usepackage{multirow}
\usepackage{makecell}
\usepackage[utf8]{inputenc}
\usepackage[T1]{fontenc}
\usepackage[english]{babel}

\def\BibTeX{{\rm B\kern-.05em{\sc i\kern-.025em b}\kern-.08em
    T\kern-.1667em\lower.7ex\hbox{E}\kern-.125emX}}

\usepackage{xurl}
\usepackage{color}
\usepackage{wrapfig}
\usepackage{textcomp}
\newif\iffigs
\figstrue 

\usepackage[fleqn]{amsmath}
\usepackage{float}
\usepackage{setspace}
\usepackage{graphicx}
\usepackage{mathrsfs}
\usepackage{amssymb,amsfonts}
\usepackage{nicefrac}
\usepackage{algorithm}
\usepackage[noend]{algpseudocode}

\makeatletter
\newcommand\fs@spaceruled{\def\@fs@cfont{\bfseries}\let\@fs@capt\floatc@ruled
  \def\@fs@pre{\vspace{0.4\baselineskip}\hrule height.8pt depth0pt \kern2pt}%
  \def\@fs@post{\vspace{-0.4\baselineskip}\kern2pt\hrule\relax\vspace{-12pt}}%
  \def\@fs@mid{\kern2pt\hrule\kern2pt}%
  \let\@fs@iftopcapt\iftrue}
\makeatother

\usepackage{listings}
\lstnewenvironment{itemlisting}[1][]
 {%
  \mbox{}
  \vspace*{-\baselineskip}
  \lstset{
    xleftmargin=\leftmargin,
    linewidth=\linewidth,
    #1
  }%
 }
 {}
\usepackage{flushend}
\usepackage{multicol}

\usepackage{lipsum}

\usepackage[pdfa,colorlinks,bookmarksopen,bookmarksnumbered,allcolors=black,urlcolor=blue]{hyperref}
\usepackage[
	activate   = {true},
	protrusion = false,
	expansion  = true,
	kerning    = true,
	spacing    = true,
	tracking   = false,
	auto       = true,
	selected   = true,
	factor     = 1000,
	stretch    = 10,
	shrink     = 10,
]{microtype}

\usepackage[nameinlink,capitalise]{cleveref}
\crefname{line}{line}{lines}
\crefname{figure}{Fig.}{Figs.}
\Crefname{figure}{Fig.}{Figs.}
\crefname{equation}{Eq.}{Eqs.}
\Crefname{equation}{Eq.}{Eqs.}
\crefname{section}{Sec.}{Secs.}
\Crefname{section}{Sec.}{Secs.}
\crefname{definition}{Def.}{Defs.}
\Crefname{definition}{Def.}{Defs.}
\crefname{algorithm}{Alg.}{Algs.}
\Crefname{algorithm}{Alg.}{Algs.}
\crefname{assumption}{Asm.}{Asms.}
\Crefname{assumption}{Asm.}{Asms.}
\crefname{subassumption}{Asm.}{Asms.}
\Crefname{subassumption}{Asm.}{Asms.}
\Crefname{problem}{Problem}{Problems}
\crefname{problem}{Problem}{Problems}
\newif\ifanon
\anontrue
\title{\LARGE \bf Real-Time, Energy-Efficient, Sampling-Based\\Optimal Control via FPGA Acceleration}

\begin{document}

\author{Tanmay Desai$^{1}$, Brian Plancher$^{2,3}$, and R. Iris Bahar$^{1}$%
\thanks{$^{1}$Colorado School of Mines {\tt\footnotesize \{desaitanmay, ririsbahar\}@mines.edu}}%
\thanks{$^{2,3}$Barnard College, Columbia University, and Dartmouth College. {\tt\footnotesize plancher@dartmouth.edu}}%
}

\maketitle

\begin{abstract}
    Autonomous mobile robots (AMRs), used for search-and-rescue and remote exploration, require fast and robust planning and control schemes. Sampling-based approaches for Model Predictive Control, especially approaches based on the Model Predictive Path Integral Control (MPPI) algorithm, have recently proven both to be highly effective for such applications and to map naturally to GPUs for hardware acceleration.
However, both GPU and CPU implementations of such algorithms can struggle to meet tight energy and latency budgets on battery-constrained AMR platforms that leverage embedded compute. 
To address this issue, we present an FPGA-optimized MPPI design that exposes fine-grained parallelism and eliminates synchronization bottlenecks via deep pipelining and parallelism across algorithmic stages.
This results in an average 3.1x to 7.5x speedup over optimized implementations on an embedded GPU and CPU, respectively, while simultaneously achieving a 2.5x to 5.4x reduction in energy usage. 
These results demonstrate that FPGA architectures are a promising direction for energy-efficient and high-performance edge robotics.

\end{abstract}


\section{Introduction}\label{sec:intro}

Autonomous Mobile Robots (AMRs) require robust sensing, planning, and control for real-world applications ranging from search-and-rescue to remote exploration and elder care~\cite{Broekens,Fragapane,krotkov2018darpa,tranzatto2022cerberus}. Recently, sampling-based stochastic optimal control algorithms—particularly Model Predictive Path Integral (MPPI) control~\cite{7487277}—have demonstrated impressive results on diverse robotic platforms~\cite{pravitra2020L1,multileg,biasedmppi,alvarez2025real,xue2025full}. These methods are amenable to GPU acceleration~\cite{articlemppi,vlahov2024mppi}, joining a growing ecosystem of GPU-accelerated robotics libraries~\cite{liang2018gpu,plancher2018performance,cervera2020gpu,makoviychuk2021isaac,miki2022elevation,sundaralingam2023curobo,huang2025prrtc}.

However, GPU-based implementations incur substantial power consumption, limiting their applicability to battery-constrained AMRs~\cite{liu2020hardware,wan2021survey,boroujerdian2022role}. Field Programmable Gate Arrays (FPGAs) offer a compelling alternative through domain-specialized hardware that eliminates GPU synchronization overhead and memory-access stalls. Yet realizing this promise requires more than implementation effort—it demands algorithmic restructuring. Naive ports of existing GPU/CPU code to HLS will not unlock FPGA potential. Instead, effective FPGA acceleration of sampling-based optimal control demands explicit hardware-software co-design: systematic analysis of algorithm structure, exposure of fine-grained parallelism, and deliberate mapping onto hardware resources under timing constraints.

Sampling-based stochastic control algorithms exhibit a common pattern: sample generation, parallel rollouts, cost evaluation, and reduction to extract optimal control. On CPUs and GPUs, these stages operate sequentially or with coarse-grained parallelism, creating implicit data dependencies and synchronization bottlenecks. FPGA acceleration requires restructuring to expose deep pipelining and parallel dataflows. This is non-trivial: (i) breaking dependencies in reduction/weighting phases, (ii) reformulating sequential operations (e.g., cost summation) as parallel trees, (iii) selecting numerical representations (fixed-point vs. floating-point), balancing precision and hardware utilization, and (iv) partitioning memory hierarchies (BRAM, URAM) to sustain parallel throughput. These are architectural imperatives—not optional tuning—that fundamentally define achievable latencies and energy profiles.

To our knowledge, this is the first systematic FPGA acceleration of sampling-based stochastic optimal control for robotics. While FPGA-accelerated control has been explored for classical methods (e.g., trajectory planning, perception), the application to stochastic sampling-based optimal control—with its unique combination of parallel Monte Carlo sampling and sequential reduction—remains unexplored. We address this gap through hardware-software co-design, instantiated on MPPI. Our approach leverages three core insights: (1) Dependency analysis characterizing parallelizable phases (sample generation, rollouts) vs. sequential phases (cost aggregation), enabling principled pipelining; (2) Dataflow-centric architecture restructuring algorithms from sequential control-flow into deeply-pipelined spatial dataflow, eliminating synchronization; (3) Resource-aware numerics co-designing fixed-point arithmetic and memory patterns to maximize hardware utilization without sacrificing accuracy.

We implement this methodology on an Xilinx ZCU102 FPGA, achieving parallel trajectory rollout generation via deep pipelining, tree-reduction-based cost accumulation, and optimized memory partitioning with comprehensive scalability analysis. Closed-loop trajectory tracking experiments demonstrate 3.1x–7.5x latency improvements and 2.2x–5.4x energy gains over optimized CPU/GPU baselines—gains from algorithmic restructuring exposing fine-grained parallelism, not hardware class differences. Our methodology generalizes across sampling-based stochastic control techniques, positioning this work as a blueprint for FPGA acceleration of a broad class of robotics algorithms.

\section{Related Work} \label{sec:related}
As previously mentioned, FPGAs have demonstrated fast, accurate, scalable, and energy-efficient computation for a variety of robotics applications, embodiments, and algorithms~\cite{murray2016robot,mcinerney2018survey,wan2021survey,plancher2021accelerating,neuman2021robomorphic,9221518, 10.1145/3400302.3415766,wan2021survey,7793243,astarfpga}. 
Within motion planning and control specifically, FPGA efforts have concentrated on search-based planning and on na\"ively parallel kernels that map well to spatial dataflow, such as collision checking and dynamics (gradient) computations~\cite{neuman2021robomorphic,plancher2021accelerating,murray2016robot,wan2021survey}, as well as control of \emph{deterministic} linear systems~\cite{mcinerney2018survey} and their applications to robotics, power electronics, and fusion~\cite{suiExploringPotentialFPGA2025,luciaDeepLearningBasedModel2021,dong_standoff_2023,gerksicFinitewordlengthFPGAImplementation2021,hartley_predictive_2013,liFPGAAcceleratedModel2022}. A complementary line of work has also leveraged FPGAs for model inference of offline learned control laws, trading exact constraint handling for speed~\cite{luciaDeepLearningBasedModel2021,dong_standoff_2023,suiExploringPotentialFPGA2025}. Importantly, across all of this work, stochastic optimal control remains underexplored on FPGAs. This paper fills that gap with a pipelined, dataflow FPGA design that exposes fine-grained parallelism across the full MPPI pipeline.

\section{Background}\label{sec:background}
The Model Predictive Path Integral Control (MPPI) algorithm (Algorithm~\ref{alg:mppi}) is a sampling-based stochastic optimal control method that considers a stochastic dynamical system with the state and controls at time $t$ denoted as $x_{t}\in \mathbb{R}^n$ and $u_t\in \mathbb{R}^m$ and the stochastic disturbance denoted as $w_t\sim {N}\left(0, \Sigma_u\right)$. Under a dynamics model, $x_{t+1} = f(x_t, u_t, w_t)$ and timestep $\Delta t$, the discrete time Euler integrator for a batch of $K$ stochastic trajectories under the same control sequence can be defined as:
\begin{equation}\label{eq:integrator}
    x_{t+1}^{(k)}=x_t^{(k)}+f\left(x_t^{(k)}, u_t+w_{t}^{(k)}\right) \cdot \Delta t.
\end{equation}
The MPPI algorithm proceeds by rolling out these $K$ trajectories and then updating the nominal control sequence through a weighting of the resulting trajectory costs, $J$: 
\begin{equation}\label{eq:stage-total}
    J^{(k)}=\phi\left(x_N^{(k)}\right) + \sum_{t=0}^{N-1} c_{t}^{(k)}\left(x_t^{(k)},u_t,w_t^{(k)}\right).
\end{equation}
Commonly, the stage cost $c$, and terminal cost $\phi$, are formed as the sum of a tracking cost on the deviation of the state from the reference trajectory as well as a total control input cost, weighted by matrices $Q\in \mathbb{R}^{n \times n}$ and $R\in \mathbb{R}^{m \times m}$:
\begin{align}\label{eq:stage}
    c_{t}^{(k)} &= \left(x_t^{(k)} - x_{t}^{\mathrm{ref}}\right)^\top Q \left(x_t^{(k)} - x_{t}^{\mathrm{ref}}\right) \nonumber \\
           &\quad + \left(u_t + w_t^{(k)}\right)^\top R \left(u_t + w_t^{(k)}\right),
\end{align}
\begin{equation}\label{eq:terminal}
    \phi_k=\left(x_N^{(k)}-x_{N}^{\mathrm{ref}}\right)^{\top} Q_f\left(x_N^{(k)}-x_{N}^{\mathrm{ref}}\right).
\end{equation}

MPPI then updates the control sequence by favoring low-cost roll-outs through an exponential weighting:
\begin{equation}\label{eq:control-update}
u_t \leftarrow u_t + \sum_{k=0}^{K-1} \alpha_k w_{t}^{(k)}, \quad  \alpha_k = \frac{\exp\left(-\frac{1}{\lambda}J_k\right)}{\sum_{j=0}^{K-1} \exp\left(-\frac{1}{\lambda}J_j\right)}.
\end{equation}
Here, the temperature hyperparameter $\lambda>0$ controls the balance between exploration and exploitation as a smaller $\lambda$ concentrates weight on lower-cost trajectories.

When used for online model predictive control (MPC), this process repeats at each control step until convergence or a maximum number of iterations is reached. The first control, $u_0$, is then applied to the real system, at which point the new system state $x_0$ is determined from sensor feedback, and the process repeats until the target is reached.

We note that as a $0^{th}$-order optimization method, MPPI is particularly well suited for nonlinear systems with non-convex cost functions, and its na\"ively-parallel computation structure enables easy GPU deployments for real-time planning applications. As such, many recent papers using such methods have demonstrated impressive results on varied platforms and tasks~\cite{pravitra2020L1,multileg,biasedmppi,alvarez2025real,xue2025full}.

\begin{algorithm}[!t]
\caption{Model Predictive Path Integral Control (MPPI)}
\label{alg:mppi}
\begin{algorithmic}[1]
\State \textbf{Inputs:} $x_0$, $X^{\mathrm{ref}}$, $N$, $K$, $\lambda$, $\Sigma_u$, $\text{max\_iters}$
\State \textbf{Initialize:} nominal controls $u_{0:N-1}$
\For{iteration $i = 0,\ldots,\text{max\_iters}$}
  \For{$k=0$ \textbf{to} $K-1$} 
    \State $x_0^{(k)} \gets x_0,\quad J^{(k)} \gets 0$
    \For{$t=0$ \textbf{to} $N-1$}
      \State {\color{purple} Compute $x_{t+1}^{(k)}$ via~\eqref{eq:integrator}}
      \State {\color{blue} $J^{(k)} \mathrel{+}=  c_t^{(k)}$ via~\eqref{eq:stage}}
    \EndFor
    \State {\color{blue} $J^{(k)} \mathrel{+}= \phi^{(k)}$ via~\eqref{eq:terminal}}
  \EndFor
  \For{$t=0$ \textbf{to} $N-1$}
    \State {\color{green}Compute $u_t$ via~\eqref{eq:control-update}}
  \EndFor
\EndFor
\State \Return $X,U$
\end{algorithmic}
\end{algorithm}

\section{MPPI Computational Dataflow Analysis}\label{sec:algo}
As described in Section~\ref{sec:background}, MPPI comprises four main computational operations per control update:
(i) \emph{Sampling random noise} for the control perturbations,
(ii) \emph{Dynamics rollouts} to propagate states (Eq.~\eqref{eq:integrator}),
(iii) \emph{Stage and terminal cost evaluation} per each state along each rollout (Eqs.~\eqref{eq:stage},~\eqref{eq:terminal}), and
(iv) \emph{Exponential weighting and control updates} (Eq.~\eqref{eq:control-update}).

From these computational operations emerge recurring computation patterns:
(i) \emph{coarse-grained parallelism} (between trajectories),
(ii) \emph{fine-grained parallelism} (small matrix operations inside rollouts and costs),
(iii) \emph{(pseudo-)random number generation} for sampling control perturbation noise, and
(iv) \emph{global reductions} across the total costs to compute the control updates.
These operations can also be \emph{pipelined} so that different trajectories (or different timesteps of the same trajectory) reside in different operations concurrently.
In practice, the boundary between “parallel blocks” and “pipelined operations” is a software co-design knob that should be adapted to the target hardware (e.g., CPU, GPU, or FPGA) to balance resource utilization and synchronization overhead.
Table~\ref{tab:pattern-fit} summarizes how each platform maps to these computational patterns.

\begin{table}[t]
\small
\centering
\caption{Key Computational Characteristics of MPPI: Architectural Fit}
\label{tab:pattern-fit}
\begin{tabular}{lccc}
\toprule
\textbf{Pattern} & \textbf{CPU} & \textbf{GPU} & \textbf{FPGA} \\
\midrule
Coarse-grained   & Poor & \textbf{Excellent} & Good \\
parallelism (trajectories) & & & \\ \hline
Fine-grained parallelism  & Moderate & Good & \textbf{Excellent} \\
(small mat-ops) & & & \\ \hline
(Pseudo-)Random  & \textbf{Excellent} & Good & Poor \\
number generator & & & \\ \hline
Global reductions & Poor & Good & \textbf{Excellent} \\
\bottomrule
\end{tabular}
\end{table}

\textbf{CPU fit with computational patterns.}
Modern CPUs handle control flow and scalar work well, and can leverage high-quality software libraries for (Pseudo-)Random Number Generators with low latency.
However, despite recent advances in multi-threading and SIMD computations, they offer limited throughput for both coarse-grained (many trajectories) and fine-grained (per-timestep small matrix-ops) parallelism as compared to GPUs and FPGAs. Similarly, global reductions tend to be bandwidth/latency bound due to a similar lack of large-scale parallel computational resources.
As a result, for MPPI, CPUs are best as a correctness/portability baseline or when the sample count $K$ is small.

\textbf{GPU fit with computational patterns.}
GPUs excel at coarse-grained parallelism across many trajectories and provide solid support for fine-grained math via wide SIMD and fast shared memory (L1 cache).
They also offer decent (Pseudo-) Random Number Generator libraries and reasonably efficient global reductions. Nevertheless, repeated synchronizations and irregular access patterns can limit utilization and prevent ultra-performant fine-grained parallelism, and thus overall compute performance.
GPUs are a strong match when $K$ is large and kernels are uniform, but may face diminishing returns when stages introduce tight dependencies or frequent global reductions. As such, while MPPI's na\"ively-parallel rollouts work well on the GPU, the need to synchronize for global reductions for each parallel computed trajectory and the (potentially complex) nonlinear dynamics computations may not be computed with maximal efficiency.

\textbf{FPGA fit with computational patterns.}
FPGAs enable custom dataflow pipelines that exploit fine-grained parallelism extremely well and implement deterministic, low-latency reduction trees for global operations. However, they typically do not offer as many on-chip parallel compute units to scale to as large of operation as GPUs. As such, leveraging deep pipelining over custom functional units is critical for performance.
In addition, on-fabric (Pseudo-)Random Number Generators are not native to most FPGAs and require extra logic and space for implementation, typically trailing CPUs/GPUs in statistical quality~\cite{10.1145/1508128.1508139}. 

\section{MPPI FPGA Co-Design}\label{sec:method}
\begin{figure*}[!t]
    \centering
    \includegraphics[width=17.7 cm]{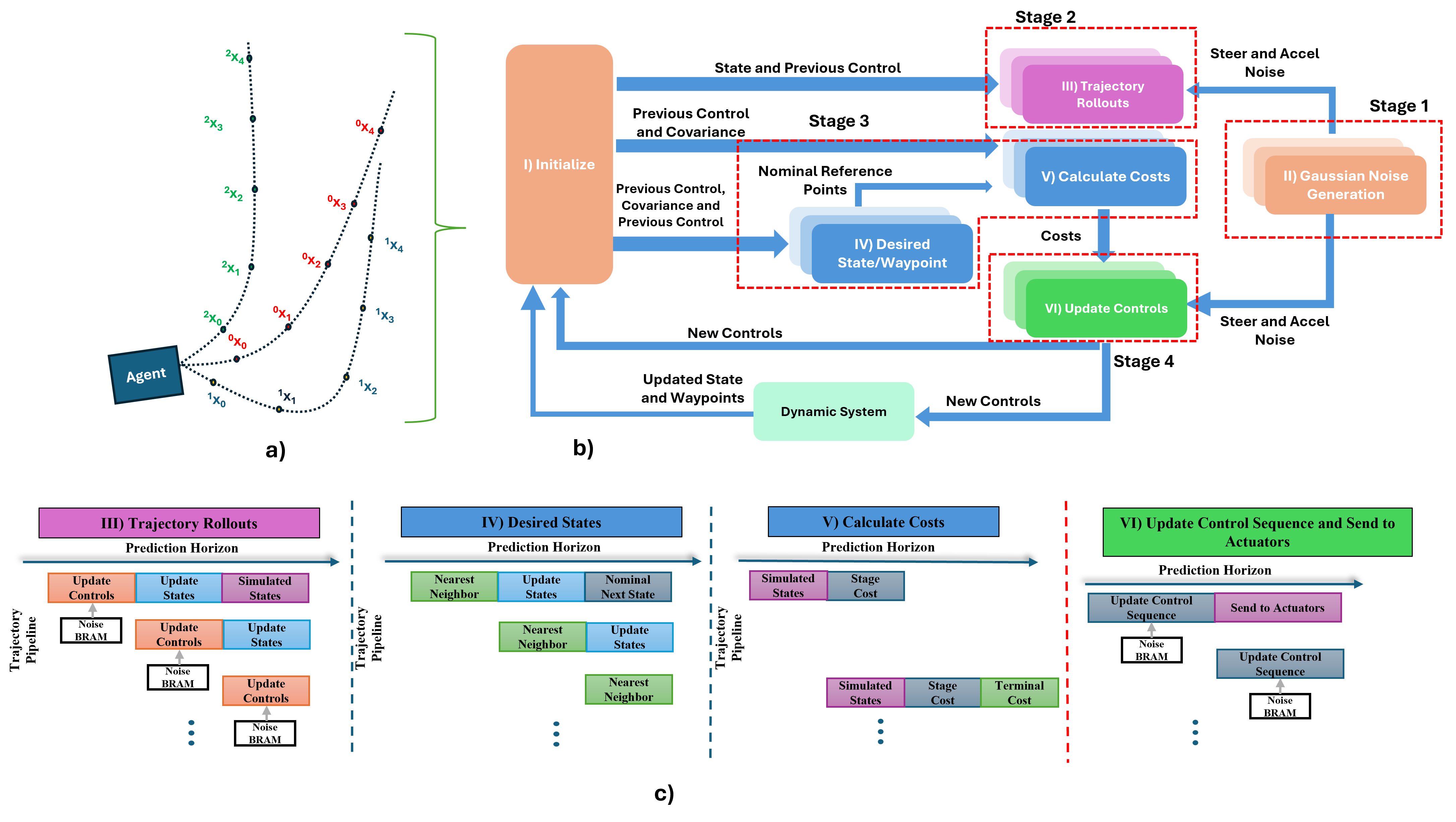}
    \vspace{-12pt}
    \caption{This figure presents the overall FPGA architecture for the MPPI algorithm. \textbf{a)} illustrates the agent's rollouts, showing the states propagated over the prediction horizon for a single control time step. \textbf{b)} depicts the 4 key stages of the MPPI control update mapped to a dataflow architecture, highlighting the parallel and overlapping execution of these stages within one iteration. \textbf{c)} details the internal functionalities of each stage from (b), further decomposed into deep pipelined sub-stages to minimize latency and maximize throughput. }
    \label{fig:stages}
    \vspace{-20pt}
\end{figure*}

To effectively co-design an FPGA implementation of MPPI, it is essential to develop a formulation that plays to the strengths and minimizes the weaknesses of FPGAs in the context of the MPPI algorithm, as discussed in Section~\ref{sec:algo}. As such, and as shown in Fig.~\ref{fig:stages}, our overall design maximizes the excellent fine-grained parallelism and global reduction capabilities on the FPGA to enable fast and efficient overall computation. 
The following subsections provide an in-depth examination of the design process pertinent to the MPPI architecture. We first build on the prior section, which characterizes data dependencies, and extend that to understand per-stage throughput on the FPGA. We then use this analysis to align producer–consumer rates across our computational pipeline. This drives scheduling, buffer depths, and data representations that maximize FPGA utilization while meeting latency and energy goals.

\subsection{Overall Design}
We detail the overall final design in Fig.~\ref{fig:stages}. Here, Fig.~\ref{fig:stages}(b), in the upper right, shows the decomposition of a single control iteration into 4 constituent computational stages, where each layer represents parallel execution units that exhibit producer-consumer concurrency through pipelined data flow. This constitutes the coarse-grained parallelism achieved through task-level dataflow decomposition of the MPPI algorithm due to the many naturally parallel trajectories. Fig.~\ref{fig:stages}(a), in the upper left, visualizes this parallelism for a sample size of $K=3$ trajectories, where $^k{x_t}$ denotes the state reached by trajectory $k\in[0, K-1]$ at a time step $t\in[0, N-1]$ within the planning horizon. Finally, in Fig.~\ref{fig:stages}(c), in the bottom, we demonstrate that each computational stage from (b) encompasses internal operations that are further parallelized through fine-grained pipelining, exploiting instruction-level and operation-level parallelism within individual algorithmic components.

As mentioned in Sec.~\ref {sec:algo}, the FPGA fabric provides a fixed pool of parallel compute units, which we keep busy via custom dataflows; this makes deeply pipelined architectures particularly powerful. Accordingly, we break MPPI into a four-stage pipelined dataflow: (1) Gaussian noise generation, (2) trajectory rollouts, (3) cost calculations, and (4) weighting and control sequence updates. Stages 1–3 admit substantial coarse-grained parallelism across the $K$ trajectories and fine-grained parallelism within the computations in each stage, while Stage 4 is a global reduction. We exploit this structure on the FPGA to overlap stages and maximize utilization.

\subsection{Stage 1: Gaussian Noise Generation}
As noted in Sec.~\ref{sec:algo}, while CPUs and GPUs offer built-in random number generators, FPGAs do not. This is therefore a critical stage in designing an algorithm to ensure optimal performance, given its reliance on randomness. To address this, we considered multiple pseudo-random number generators (PRNGs). The end end our finalists were two classes of PRNGs, both common and popular in FPGA implementations, derived from both linear feedback shift registers (LFSRs)~\cite{8073966} and XOR-Shift~\cite{xorshift} operations. 
Ultimately, we selected an XOR-Shift-based scheme as it was able to produce a full 32-bit result every clock cycle, unlike LFSR, which produces one bit at a time. This makes XOR-Shift immediately suitable for high-throughput noise generation without the heavy resource cost required to parallelize LFSR. 

Unfortunately, both LFSR and XOR-Shift produce uniform random numbers, which are unsuitable for our application, since MPPI requires Gaussian-distributed noise. To address this, we use the Box-Muller transform (Eq.~\ref{eq:boxmuller}) to convert uniform samples into standard normal (Gaussian) noise. 

\begin{equation}\label{eq:boxmuller}
    \begin{gathered}
R_i = \sqrt{-2 \ln \left(\tilde{U}_{1, i}\right)} \\
\Theta_i = 2 \pi \tilde{U}_{2, i} \\
\text{steer\_samples}[i] = R_i \sin \left(\Theta_i\right) \\
\text{accel\_samples}[i] = R_i \cos \left(\Theta_i\right)
\end{gathered}
\end{equation}

\begin{figure}[bth]
    \centering
    \includegraphics[width=\linewidth]{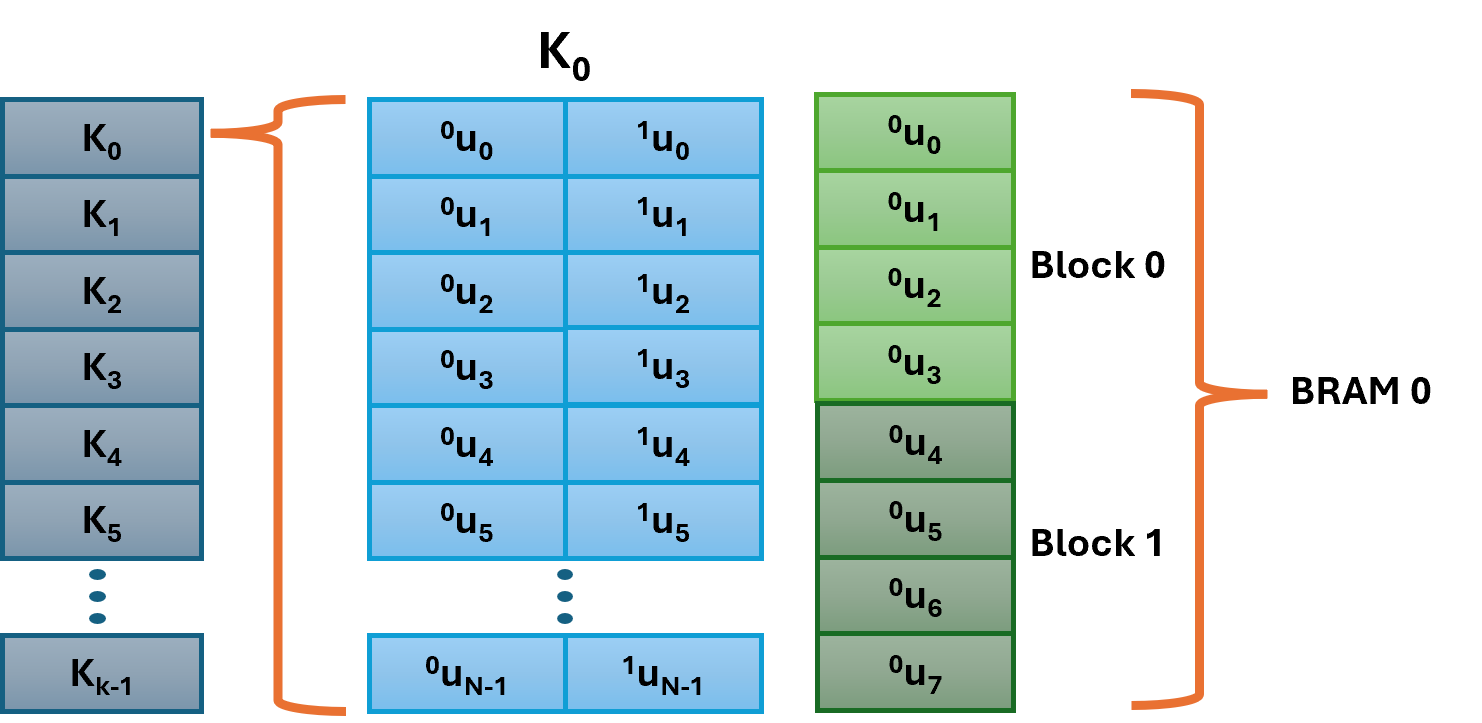}
    \caption{Array partitioning:  Noise values for control arrays are partitioned into 4 sub-blocks.  As shown, there are $K$ trajectories, where each trajectory produces $N$ pairs of noise values (for acceleration and steering) for the $N$ time steps, which are written to BRAM in blocks of 4 at a time.}
    \label{fig:arraypart}
    \vspace{-15pt}
\end{figure}

To minimize generation latency, Gaussian noise for each control is produced in parallel every cycle, with samples written concurrently across $\mathbf{P}$ memory banks to prevent port contention, where $\mathbf{P}$ represents the parallel iterations of generation.
However, the limited number of write ports in each bank of the Block RAM (BRAM) prevents simultaneous writes of control values to a single, monolithic array. To overcome this bottleneck, we apply array partitioning: the original steer and acceleration arrays are split into multiple smaller subarrays, each mapped to its own BRAM block. Fig.~\ref{fig:arraypart} illustrates how this partitioning enables true parallel writes across all trajectory states, eliminating port contention and preserving our pipeline’s throughput. This stage also behaves as a data source, resulting in simultaneous noise generation with FPGA initialization.

While this approach is effective, it demands significant floating-point computation, branching, and large memory for tables, consuming significant FPGA resources, both in terms of Block-RAM (BRAM) memory and  Digital Signal Processing (DSP) compute units. 
To minimize this, our design adopts the COordinate Rotation Digital Computer (CORDIC) algorithm~\cite{article}, which replaces multipliers and dividers with shift-and-add operations. This technique is highly amenable to pipelined FPGA implementation, with each iteration forming a lightweight pipeline stage that uses only a single adder or subtractor, enabling resource sharing and efficient computation of trigonometric, logarithmic, and exponential functions.

\subsection{Stage 2: Trajectory Rollouts}
This phase employs a dataflow architecture, leveraging parallelism across $K$ trajectories and $N$ time steps. 
The dataflow for this phase leverages 
an outer loop iterating over $K$ trajectories, partially unrolled with factor $P$, and an inner loop spanning $N$ time steps, entirely pipelined.


The data flow graph shown in Fig.~\ref{fig:stages}(c) shows the resulting fully pipelined data movement: as soon as one stage produces a result, the next stage can consume it, constrained only by data dependencies. Point-to-point wiring eliminates instruction-fetch overhead, allowing each stage to process incoming data without delay. Because this streaming transfer occurs on every clock cycle, end-to-end pipeline latency is minimized. Use of FIFO buffers between stages ensures sustained throughput by decoupling producer and consumer blocks, preventing stalls even under variable processing times. 

We also exploit fine grained parallelism in the underlying dynamics functions, Eq.~\ref{eq:integrator}, parallelizing updates across the state and control dimmensions.



\subsection{Stage 3: Cost Calculations}
The cost stage uses a streaming dataflow that reduces each trajectory to a scalar: stage/terminal costs, (Eqs.~\ref{eq:stage} and~\ref{eq:terminal}), are run in a pipelined kernel. During computation, array partitioning and per-trajectory accumulators break loop-carried dependencies, quadratic costs are inlined to cut call overhead and improve scheduling, and small matrix multiplies map to DSP blocks to increase parallelism without memory contention, fully exploiting fine-grained parallelism. As with stage 2, outputs from the $P$ parallel channels are buffered in FIFOs for downstream use.

\subsection{Stage 4: Weighting and Control Sequence Updates}
This final stage transforms trajectory cost values into an optimal control policy update through importance-weighted parameter estimation, representing the computational convergence point of the entire dataflow architecture through a large-scale reduction. 
The importance-weighted reduction implements a $K$-to-$N$ many-to-few reduction pattern, collapsing the ensemble dimension while preserving temporal structure. The temporal smoothing filter maintains causal structure through incremental updates, avoiding full-window recomputation at each time step. The refined control sequence feeds forward to the initialization stage, establishing the temporal feedback loop that enables iterative policy improvement.

\section{Experiments and Result}\label{sec:experiments}
In this section, we provide three sets of results for a racecar path tracking task, adapted from~\cite{Vlahov2024MPPIGenericAC}, describing the latency and energy reductions, as well as end-to-end tracking and success rate improvements delivered by our FPGA design. 

\subsection{Methodology}
Our FPGA design is implemented with Vitis HLS~\cite{coussy2010high} and runs on a 200 MHz Xilinx UltraScale+ MPSoC ZCU102. We compare our implementation to the state-of-the-art benchmark MPPI-generic implementation~\cite{Vlahov2024MPPIGenericAC} running on a 1020 MHz, 1024-CUDA-core NVIDIA Jetson Orin Nano 8 GB GPU. 
Existing CPU implementations of the MPPI algorithm exist in 
Python~\cite{aoki2025simplemppi} and Jax~\cite{lehtomaa2025jaxmppi}, but have known overheads.  Instead, 
we implemented our own CPU baseline using Eigen~\cite{eigenweb} for efficient linear algebra and OpenMP~\cite {dagum1998openmp} for multi-threading. Our CPU implementation was run on the 6-core Arm Cortex-A78 64-bit CPU, which is also on the NVIDIA Jetson Orin Nano 8 GB. 

\begin{figure}[!t]
    \centering
    \begin{minipage}[b]{0.47\linewidth}
        \centering
        \includegraphics[width=\linewidth]{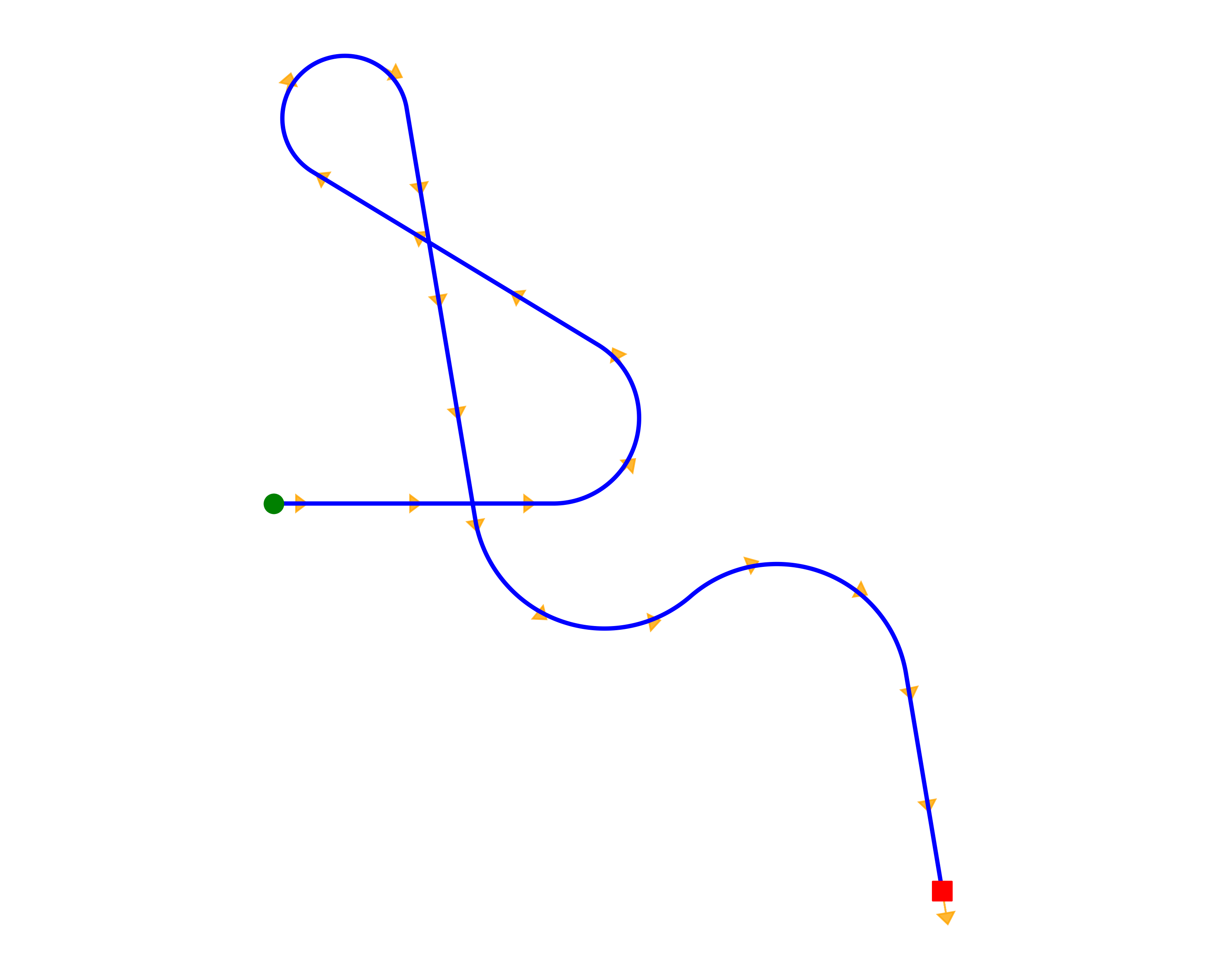}
        \subcaption{Track 1}
    \end{minipage}
    \begin{minipage}[b]{0.47\linewidth}
        \centering
        \includegraphics[width=\linewidth]{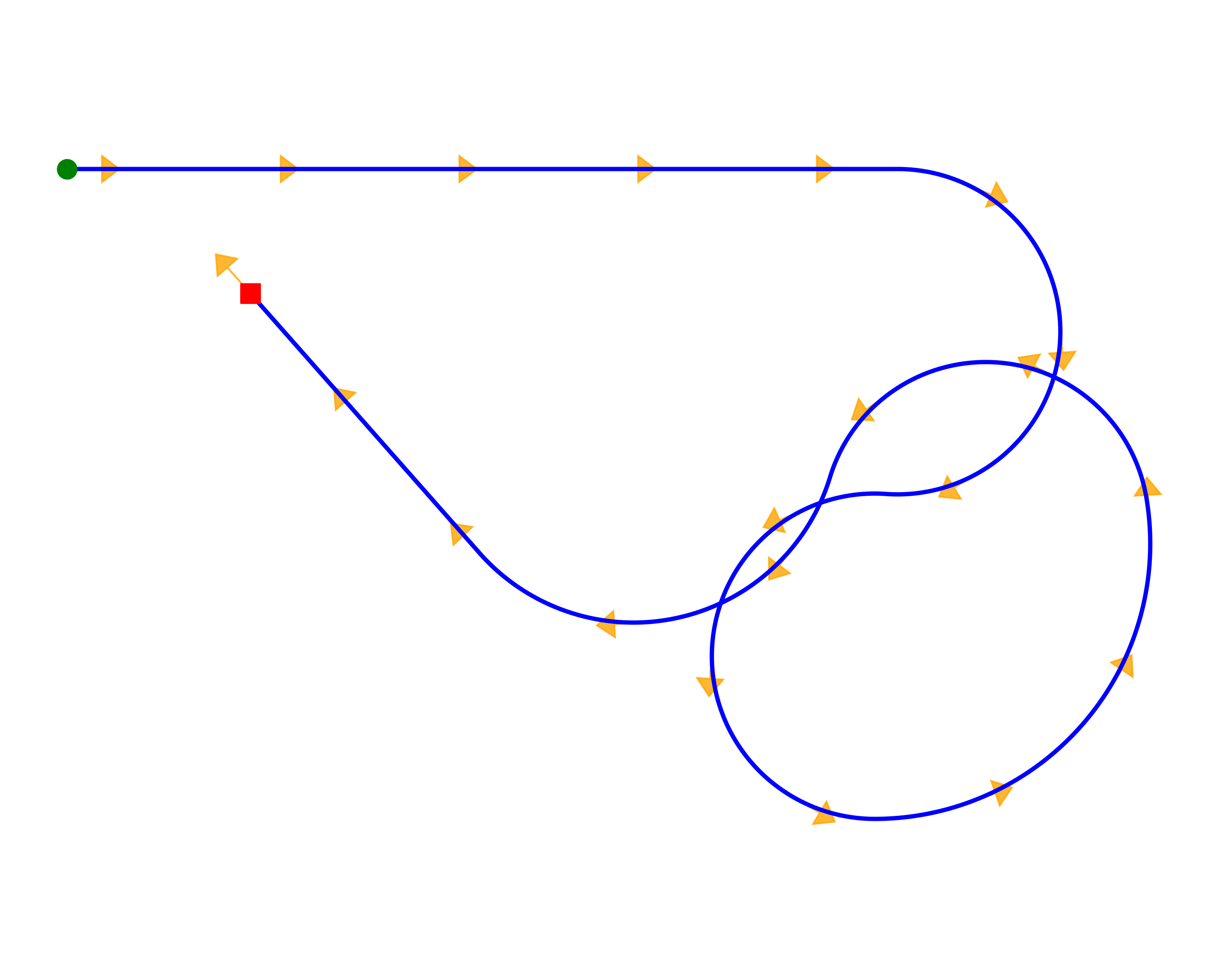}
        \subcaption{Track 2}
    \end{minipage}\\
    \begin{minipage}[b]{0.47\linewidth}
        \centering
        \includegraphics[width=\linewidth]{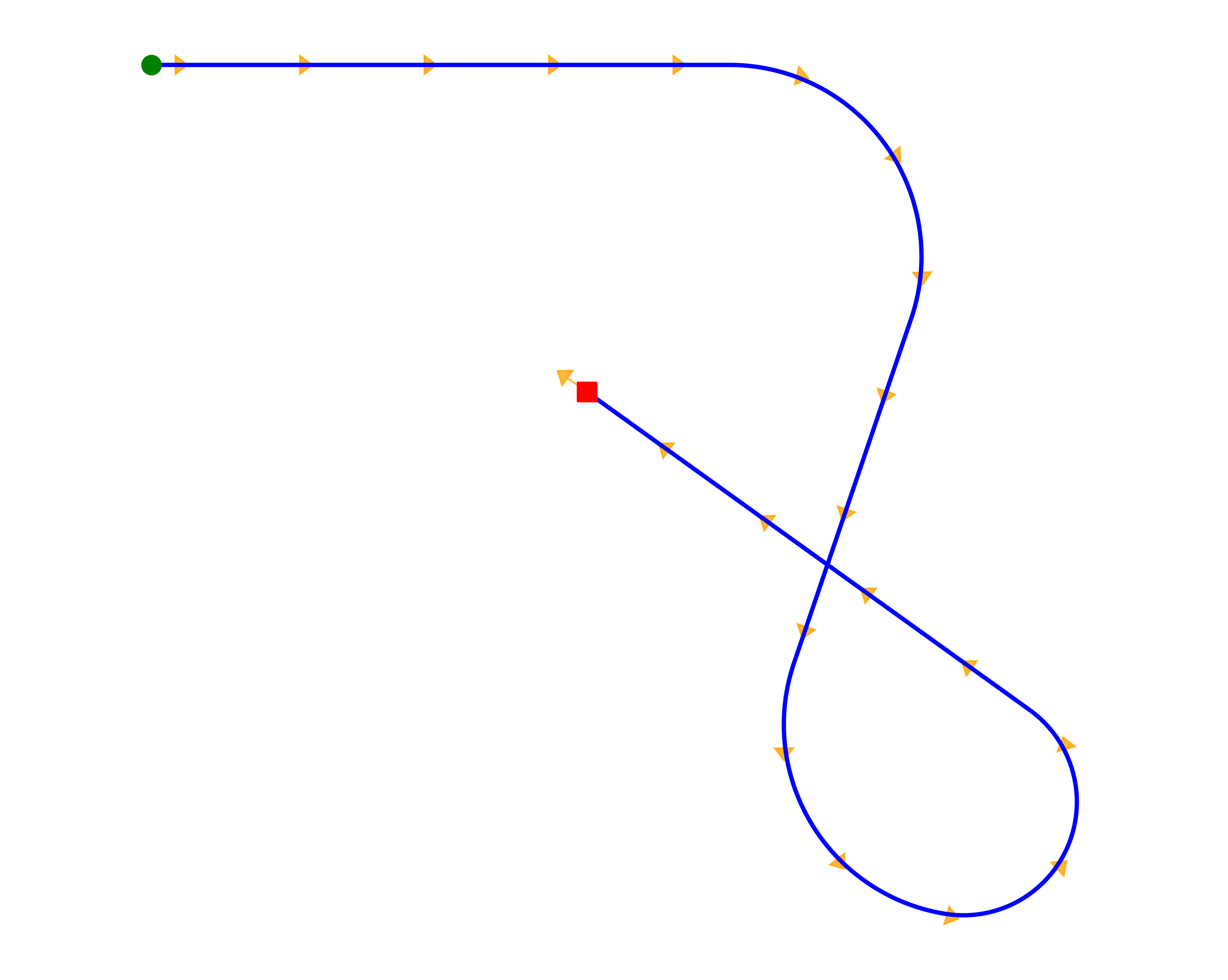}
        \subcaption{Track 3}
    \end{minipage}
    \begin{minipage}[b]{0.47\linewidth}
        \centering
        \includegraphics[width=\linewidth]{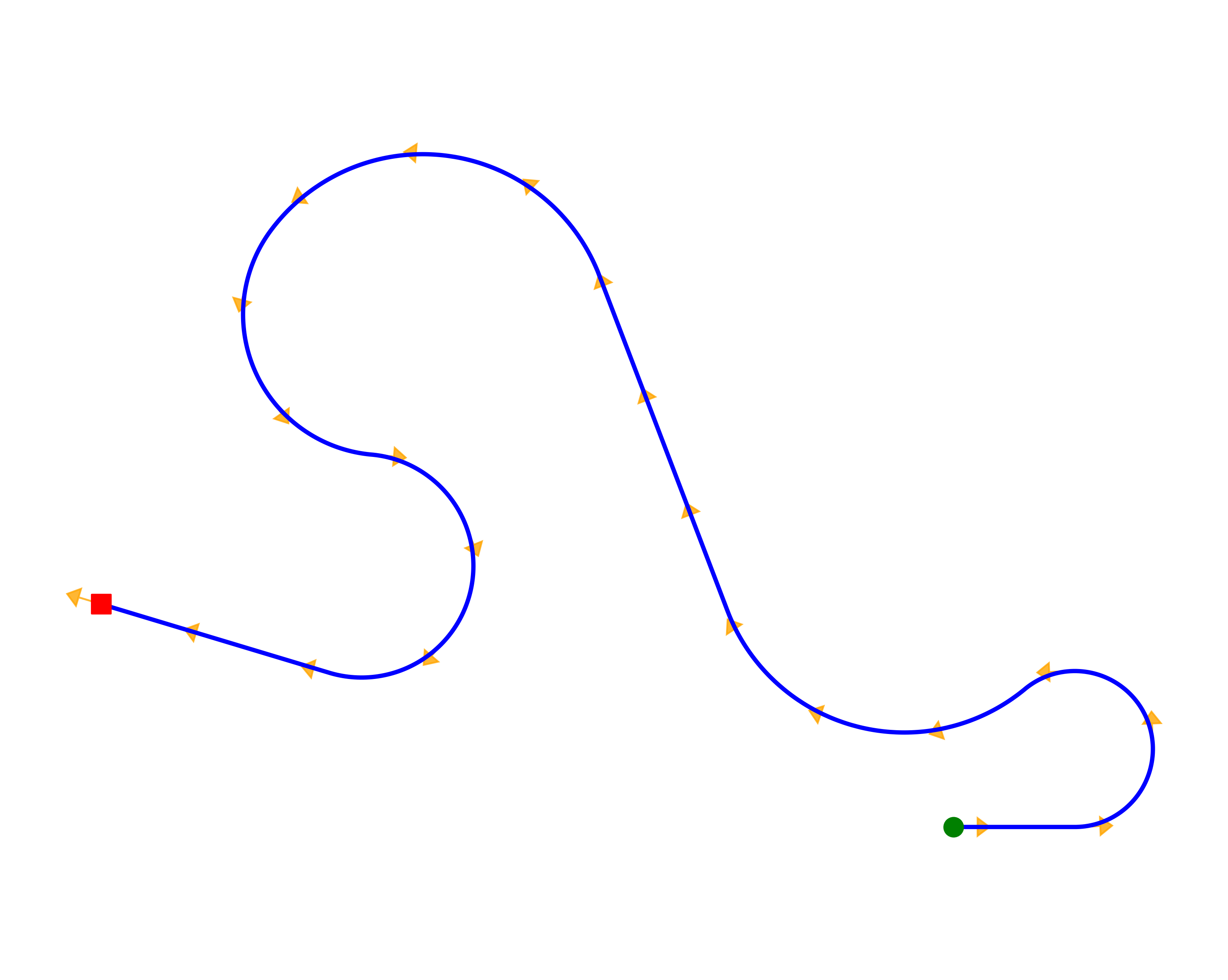}
        \subcaption{Track 4}
    \end{minipage}\\
    \begin{minipage}[b]{0.5\linewidth}
        \centering
        \includegraphics[width=\linewidth]{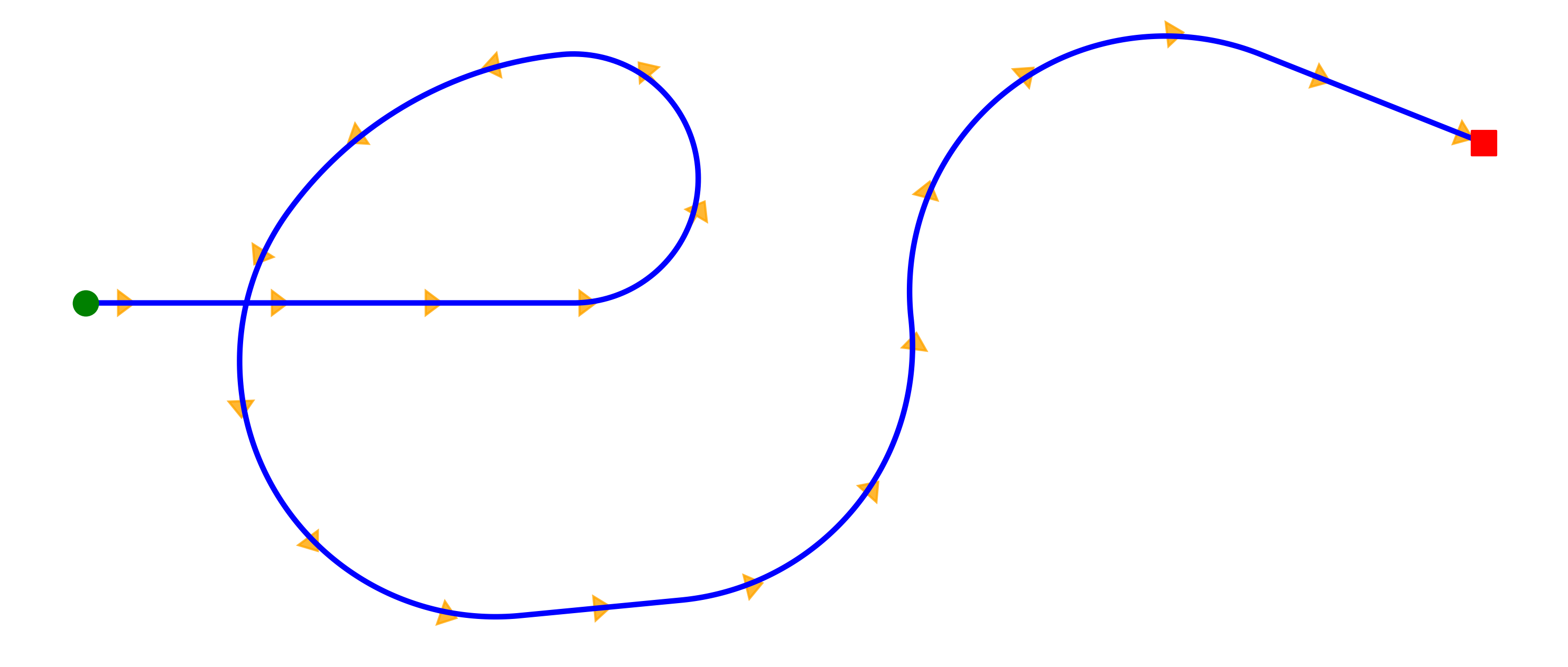}
        \subcaption{Track 5}
    \end{minipage}
    \caption{Different tracks used for our experiments.}
    \label{fig:Paths}
\end{figure}

Our experiments involved five randomly generated tracks (see Fig.~\ref{fig:Paths}), each starting from 11 different initial points, resulting in 55 total tests per device. Each test was run 20 times, and averages are reported. For each test, we measured energy usage, control-loop delay, and success rates. Consistent hyperparameters were maintained across all devices. 
All of our examples use a car with the Kinematic Bicycle model and the quadratic cost function as in~\cite{Vlahov2024MPPIGenericAC}.


\begin{table*}[!t]
    \centering
    \caption{Latency, Energy Consumption, Energy-Delay-Product (EDP), End-to-End Tracking Error, and Success Rate for the CPU, GPU, and FPGA across all experiments. Improvements and reductions are noted for the FPGA as compared to the CPU and GPU baselines, showing the FPGA's best overall performance across the board.}
    \label{tab:result_compare}
    \begin{tabular}{|c||c|c|c|c|c||c|c|c|}
    \hline \textbf{Device} & \textbf{ Timing} &\textbf{Energy} &\textbf{EDP} &\textbf{Avg Error} &\textbf{Success Rate} &\textbf{Speedup} &\textbf{Energy} &\textbf{EDP} \\
    & (ms) & (mJ) & (ms$\cdot$mJ) & (cm) & & & \textbf{Reduction} & \textbf{Gain}\\
    \hline \text{CPU} & 17.37 & 80.16 & 1392.38 & 61.6 & 93\% & 7.5x & 5.4x & 40.1x \\
    \hline \text{GPU} & 7.24 & 37.44 & 271.06 & 48.6 & 95\% & 3.1x & 2.5x & 7.8x \\
    \hline \text{FPGA \textbf{(Ours)}} & $\mathbf{2.33}$ & $\mathbf{14.90}$ & $\mathbf{34.72}$ & $\mathbf{47.6}$ & 96\% & - & - & - \\
    \hline
    \end{tabular}
\end{table*}
   
\begin{figure*}[!t]
  \centering
  \begin{subfigure}[t]{0.505\textwidth}
    \centering
    \includegraphics[width=\linewidth]{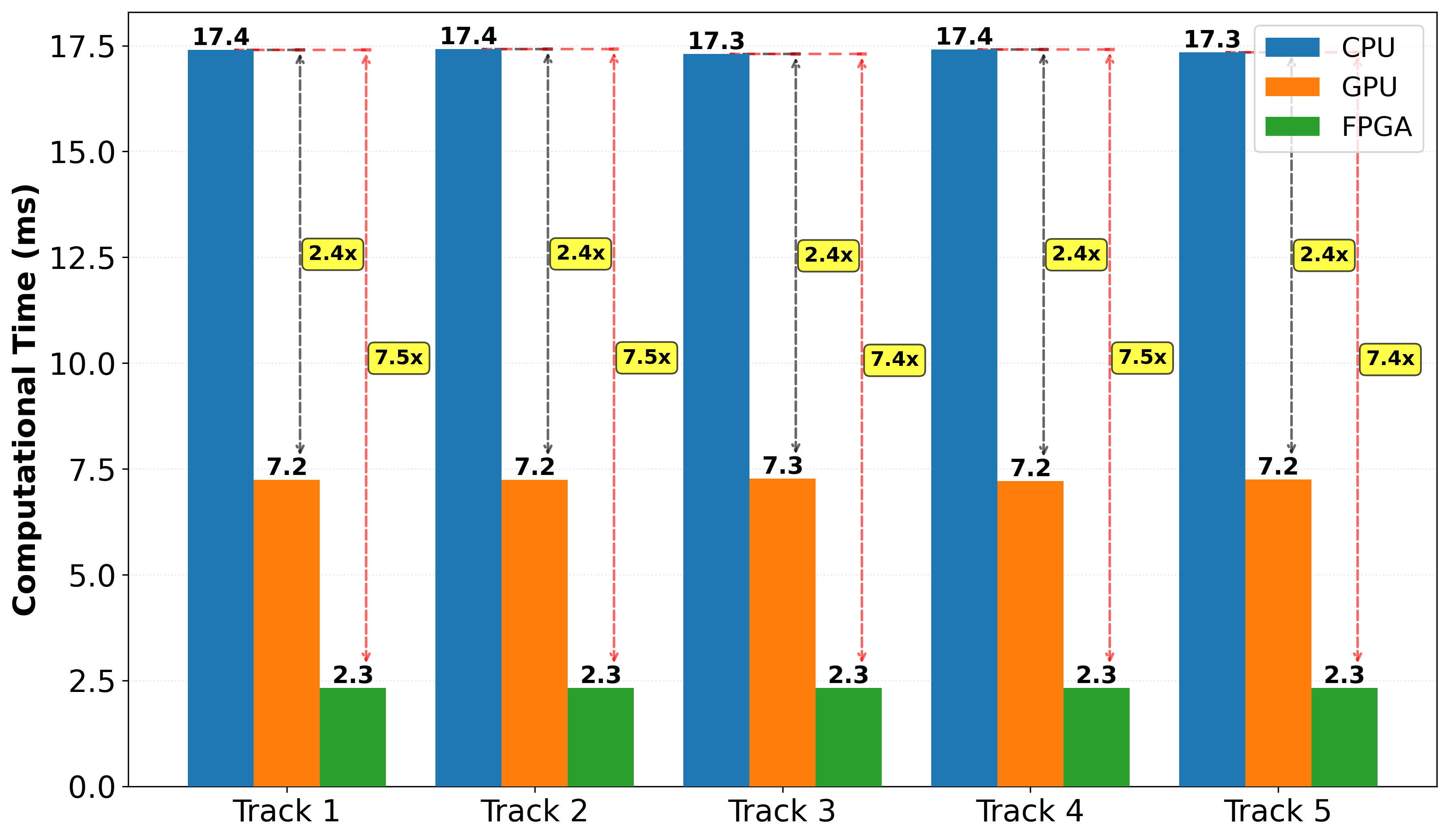}
    \caption{Average control-step latency per track for CPU, GPU, and FPGA; FPGA yields large speedups over both.}
    \label{fig:Compute}
  \end{subfigure}\hfill
  \begin{subfigure}[t]{0.47\textwidth}
    \centering
    \includegraphics[width=\linewidth]{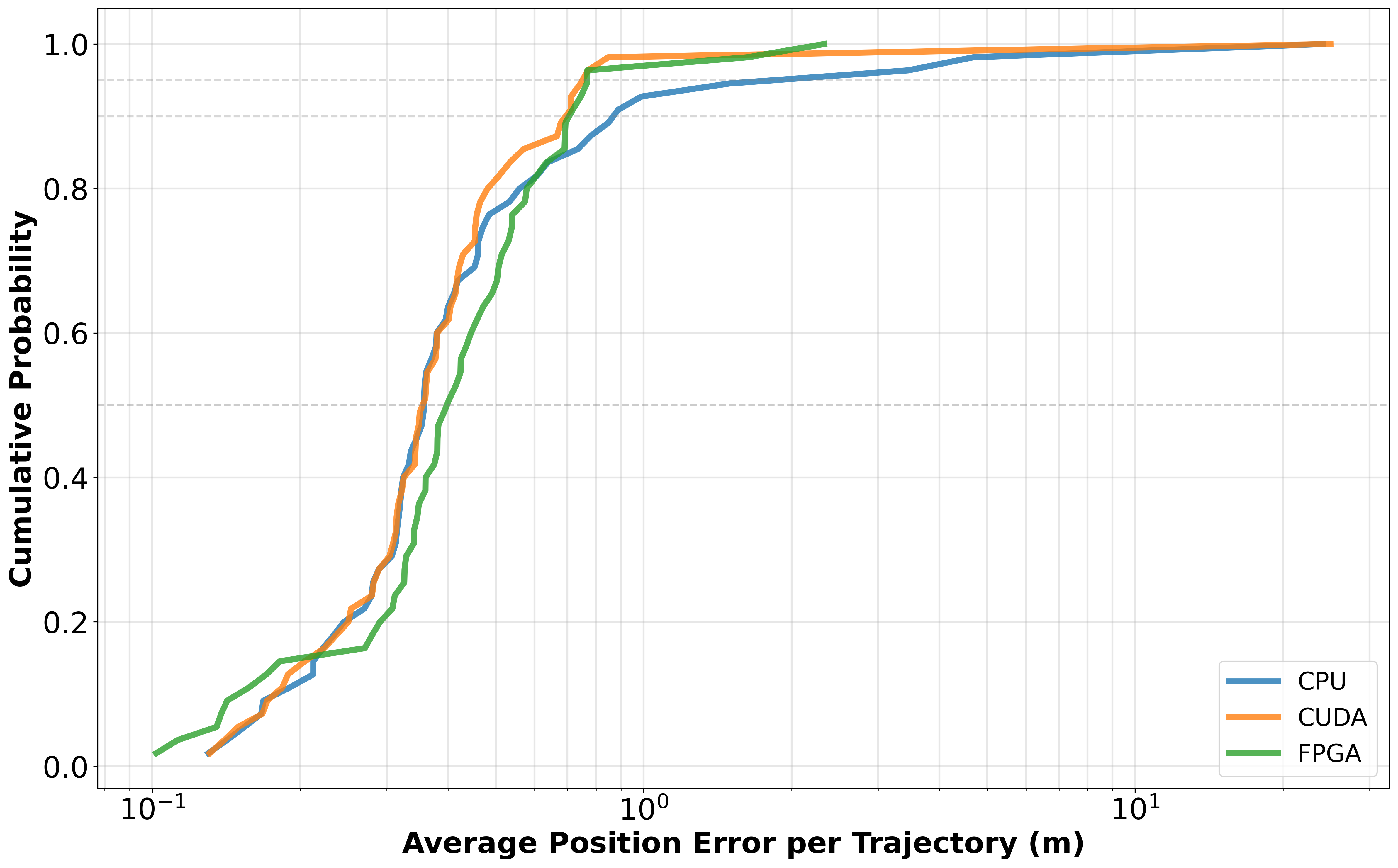}
    \caption{CDF of average position error (log scale), showing consistently low error for the FPGA.}
    \label{fig:avgerror}
  \end{subfigure}
  \caption{Latency and accuracy comparison across platforms.}
  \label{fig:latency-accuracy}
\end{figure*}

\subsection{Latency Reduction}

The average per-step computation time, defined as the duration required to execute one full control loop and generate the optimal control update, is summarized in Table~\ref{tab:result_compare} and Fig.~\ref{fig:Compute}. Our FPGA-based implementation achieves an average runtime of 2.33 ms per control step across all tracks, outperforming the NVIDIA Jetson platform by 3.1x (7.241 ms) and the multi-threaded CPU implementation by 7x (17.374 ms). These results demonstrate that the fine-grained parallelism and pipelined dataflow architecture of the FPGA delivers substantially lower latency for our target application. 

\subsection{Energy Reduction}
Table~\ref{tab:result_compare} also reports the mean energy drawn per control‐loop iteration for each platform. Our FPGA implementation consumes roughly half the energy of the Jetson Orin Nano, attributable to its bespoke dataflow pipelines and fine‐grained parallelism that eliminate the overhead of general‐purpose cores and GPU SM schedulers. Compared to the multi‐threaded CPU baseline, the FPGA  implementation has a 5.4x energy advantage, thanks in part to its clock‐gated, customized logic that activates only the necessary resources for each arithmetic operation. 
The combination of the speedups presented in the prior section and the energy reductions presented here leads to large gains in overall Energy-Delay Product (EDP) with over 40x improvements for the FPGA over the CPU and 7.8x over the GPU.

\subsection{End-to-End Tracking Performance}

Table~\ref{tab:result_compare} quantifies the average tracking error and success rate across all tasks, and Fig.~\ref{fig:avgerror} shows the cumulative density function of the average tracking error across all tests. We find that our FPGA-based implementation achieves the lowest mean deviations from the reference path. The FPGA is also the only hardware platform to not have a high error spike on any track (Fig.~\ref{fig:avgerror}), showing that despite relying on lightweight, hardware-friendly pseudo-random number generators, rather than more accurate software libraries, the FPGA’s low latency enables it to consistently provide low tracking error and high success.

\section{Conclusion and Future Work}\label{sec:conclusion}
In this work, we analyze the computational patterns present in the popular MPPI algorithm and co-design an optimized FPGA implementation. Our design 
optimizing for dataflow parallelism and avoiding control-flow checks, enabling seamless fine-grained parallel pipelining. 
Our results show an average 3.1x to 7.5x latency improvement over optimized implementations on an embedded GPU and CPU, respectively, while simultaneously achieving a 2.5x to 5.4x improvement in energy efficiency. 

There are many exciting directions for future research, including exploring fixed-point arithmetic optimization and dynamic partial reconfiguration to enable even higher performance and lower energy consumption across more complex and changing environments. We would also be interested in exploring deployments to physical robots. Fixed-point representations can reduce resource use and power consumption, but need careful precision analysis. Dynamic partial reconfiguration allows runtime adaptation in response to changing conditions, crucial for long-term autonomous missions needing recovery and adaptability.

\bibliographystyle{inc/IEEEtran}
\bibliography{inc/refs}
    
\end{document}